\documentclass[letterpaper, 10 pt, conference]{ieeeconf}  

\IEEEoverridecommandlockouts                              

\usepackage{amsmath} 
\usepackage{amssymb}  
\usepackage{graphicx}

\title{\LARGE \bf
Robot--Task Suitability in Heterogeneous Multi-Robot Task Allocation
}

\author{Daniel Weiner$^{1}$ and Raj Korpan$^{2}$%
\thanks{$^{1}$Daniel Weiner is with The Graduate Center, City University of New York (CUNY), New York, NY, USA
        {\tt\small dweiner@gradcenter.cuny.edu}}%
\thanks{$^{2}$Raj Korpan is with Hunter College and The Graduate Center, City University of New York (CUNY), New York, NY, USA
        {\tt\small raj.korpan@hunter.cuny.edu}}%
}

\begin{document}

\maketitle
\thispagestyle{empty}
\pagestyle{empty}

\begin{abstract} Heterogeneous multi-robot task allocation depends on estimates of how suitable individual robots are for candidate tasks, yet suitability can be defined in substantially different ways. We compare ten deterministic methods and two LLM-based variants spanning rule-based, similarity-based, requirement-oriented, fuzzy, and semantic approaches. Using common robot and task profiles, we compare how the methods score and rank candidate robots, identify feasibility, and select a top robot, while examining how these outcomes change with team composition, scale, spatial context, capability uncertainty, and task representation. The methods frequently disagree despite operating on identical scenarios. Generic similarity measures can violate properties expected under a one-sided requirement-satisfaction interpretation of suitability, while LLM-based scoring introduces additional stochastic and linguistic sensitivity. These results show that robot--task suitability should be treated as an explicit modeling choice whose assumptions, robustness, and decision-level consequences should be considered before allocation. \end{abstract}


\section{Introduction and Related Work}
\label{sec:introduction}

Multi-robot task allocation (MRTA) concerns assigning tasks to robots subject to constraints on resources, timing, motion, and team objectives. MRTA has been studied through centralized optimization, market- and auction-based methods, distributed coordination, heuristics, and learning-based approaches \cite{gerkey2004formal,korsah2013comprehensive,khamis2015multi,ka2024systematic}. The problem is especially challenging for heterogeneous teams, whose robots may differ in mobility, sensing, manipulation, payload, reach, endurance, communication, and environmental tolerance \cite{rizk2019cooperative}. Existing heterogeneous MRTA approaches represent these differences through capability constraints, utilities, energy-aware allocation, resilience mechanisms, or task-specific requirements \cite{ghassemi2022multi,notomista2021resilient,mayya2021resilient,calvo2025heterogeneous}. Regardless of the allocation mechanism, however, the system must determine how appropriate each robot is for each task \cite{notomista2019optimal,emam2020adaptive}. We use the term \emph{robot--task suitability} to describe how well a particular robot matches the requirements of a particular task, including considerations such as feasibility, specialization, and expected performance.

Constructing suitability values is itself a nontrivial modeling problem. Robot capabilities and task requirements may be numerical, categorical, contextual, unstructured, or uncertain, and some requirements are mandatory while others permit partial satisfaction or tradeoffs. A robot with the necessary hardware may nevertheless be a poor candidate under current conditions because of travel distance, endurance, or environmental constraints \cite{notomista2019optimal,emam2020adaptive,verma2025cf,feo2021spatially}. Prior work has considered uncertain or changing capabilities and task requirements \cite{mataric2003multi,choudhury2022dynamic}, while adaptive specialization methods model robot efficacy as initially unknown or changing during deployment \cite{emam2020adaptive,mayya2021resilient}. Reducing these factors to a scalar therefore requires assumptions about hard vs. soft constraints, capability weighting, surplus vs. deficient capability, and whether operational context is part of suitability.

Different methods encode these assumptions differently. Rule-based methods can explicitly evaluate requirements and reject infeasible pairs; vector-based approaches compare robots and tasks through geometric similarity; requirement-oriented methods measure satisfaction or deficiency; and fuzzy methods allow gradual satisfaction near numerical thresholds. Multi-objective MRTA likewise reflects the need to balance competing aspects of robot suitability \cite{tolmidis2013multi,tolmidis2017multi,wang2022cooperative,dai2023multi}. More recently, large language models (LLMs) have been used to reason over heterogeneous robot capabilities and natural-language task descriptions \cite{zhang2024context,kannan2025mtu,weiner2025hybridvoting}. Such methods offer semantic flexibility but introduce stochasticity and sensitivity to linguistic representation. These approaches encode different notions of compatibility, making their effects difficult to isolate when they are evaluated using different robot, task, and allocation models.

This work therefore examines suitability estimation as a distinct component of heterogeneous MRTA rather than proposing a new allocation algorithm. We compare ten deterministic methods, including rule-based, similarity-based, requirement-oriented, set-based, and fuzzy approaches, together with two LLM-based variants, using common robot and task profiles. We evaluate agreement and robustness under changes in team composition, scale, robot-to-task ratio, spatial scale, and capability perturbations; test five one-sided suitability properties; examine task-dependent behavior; and evaluate LLM repeatability and semantic robustness. The experiments show that suitability method affects feasibility judgments and robot rankings, that several generic similarity measures exhibit behaviors inconsistent with one-sided requirement satisfaction, and that semantic scoring introduces additional uncertainty. These findings motivate treating suitability estimation as an explicit modeling decision rather than an assumed input to downstream task allocation.


\section{Methodology and Experimental Design}
\label{sec:method}

We consider a heterogeneous robot team
$\mathcal{R}=\{r_1,\ldots,r_N\}$ and task set
$\mathcal{T}=\{t_1,\ldots,t_M\}$.
Robot profiles contain continuous attributes such as payload, reach, sensing
range, endurance, speed, processing capability, size, and location, together
with categorical attributes for mobility, sensing, manipulation,
communication, safety, environmental resistance, and special functions.
Task profiles contain corresponding requirements, navigation constraints,
duration, difficulty, priority, location, and a natural-language description.
Robots were generated from six archetypes (delivery, assembly, excavator,
bricklayer, crane, and scaffolding) and tasks from eight categories (utilities, debris removal, delivery, assembly, excavation, item elevation,
brick laying, and scaffolding). Numerical and categorical attributes were
randomized within archetypes so that robots were not
identical.

Each suitability method $m$ produces a matrix
$\mathbf{S}^{(m)}\in[0,1]^{N\times M}$, where $s_{ij}^{(m)}$ is the estimated
suitability of robot $r_i$ for task $t_j$. Ten deterministic methods were
evaluated on identical generated scenarios. The three rule-based methods
share capability categories but differ in constraint strictness:
\textit{Loose} applies soft penalties, \textit{Balanced} treats critical
payload, reach, manipulation, and battery-duration failures as infeasible,
and \textit{Strict} additionally hard-constrains communication, safety,
environmental, and navigation compatibility. \textit{Euclidean} and
\textit{cosine} similarity compare normalized robot and task representations
containing continuous, categorical, and applicable pair-dependent features.
\textit{Requirement coverage} measures the fraction of required dimensions
meeting threshold values, \textit{Jaccard similarity} measures overlap between
binarized capability representations, and \textit{fuzzy scoring} replaces
hard numerical thresholds with gradual satisfaction. \textit{Shortfall}
measures only deficits relative to requirements, while its weighted variant
assigns different importance to requirement dimensions. Thus,
similarity-based methods may respond to both surplus and deficient capability,
whereas requirement-oriented methods are primarily one-sided.

The experiments varied six problem characteristics.
\emph{Scale} varied robot and task counts over
$\{5,10,20,50,100\}$; \emph{robot-to-task ratio} varied $N/M$ over
$\{0.25,0.5,1,2,4\}$ with task count fixed. \emph{Archetype heterogeneity}
varied the robot-type pool from one to all six archetypes, while
\emph{within-archetype variation} changed the numerical and categorical
variation among robots of the same type. \emph{Spatial scale} varied map size
over $\{25,50,100,200\}$, affecting applicable distance-dependent navigation,
sensing, endurance, and proximity terms. \emph{Perturbation robustness}
modified already-generated robot profiles at magnitudes
$\{0,0.02,0.05,0.10,0.20\}$ to isolate sensitivity to profile uncertainty
from population-generation variability. We also performed five property tests that examined whether suitability increases when a required numerical capability improves, respects dominance between robots, does not decrease when task requirements are relaxed, remains unchanged when irrelevant capabilities are added, and decreases when a critical capability is removed. Agreement was also analyzed separately by task type. Unless varied experimentally, experiments used high archetype heterogeneity, within-archetype variation, and a $50\times50$ map, with each condition repeated across random seeds.

Suitability matrices $\mathbf{S}^{(m)}$ are summarized using three statistics. The \emph{mean
score} averages suitability over all robot--task pairs. The \emph{zero rate} is the fraction of pairs
assigned zero suitability and, for methods that use zero to denote
infeasibility, also measures the fraction of infeasible pairs. The \emph{mean
best score per task} averages the highest robot suitability for each task and
therefore reflects the strength of the best available match for a typical
task. Cross-method comparisons use mean absolute score difference,
matrix-level Pearson and Spearman correlation, mean per-task Spearman
correlation, exact top-robot agreement, and zero/nonzero disagreement.
Perturbation experiments additionally measure score change, correlation with
the unperturbed matrix, top-robot stability, and zero-status flips. We report
these measures separately because similarity in score magnitude, feasibility,
ranking, and selected robot need not coincide. Top-robot selection provides a
decision-level proxy for the effect of suitability on allocation, not a
complete multi-task allocation.

\begin{figure}[t]
    \centering
    \includegraphics[width=\linewidth]{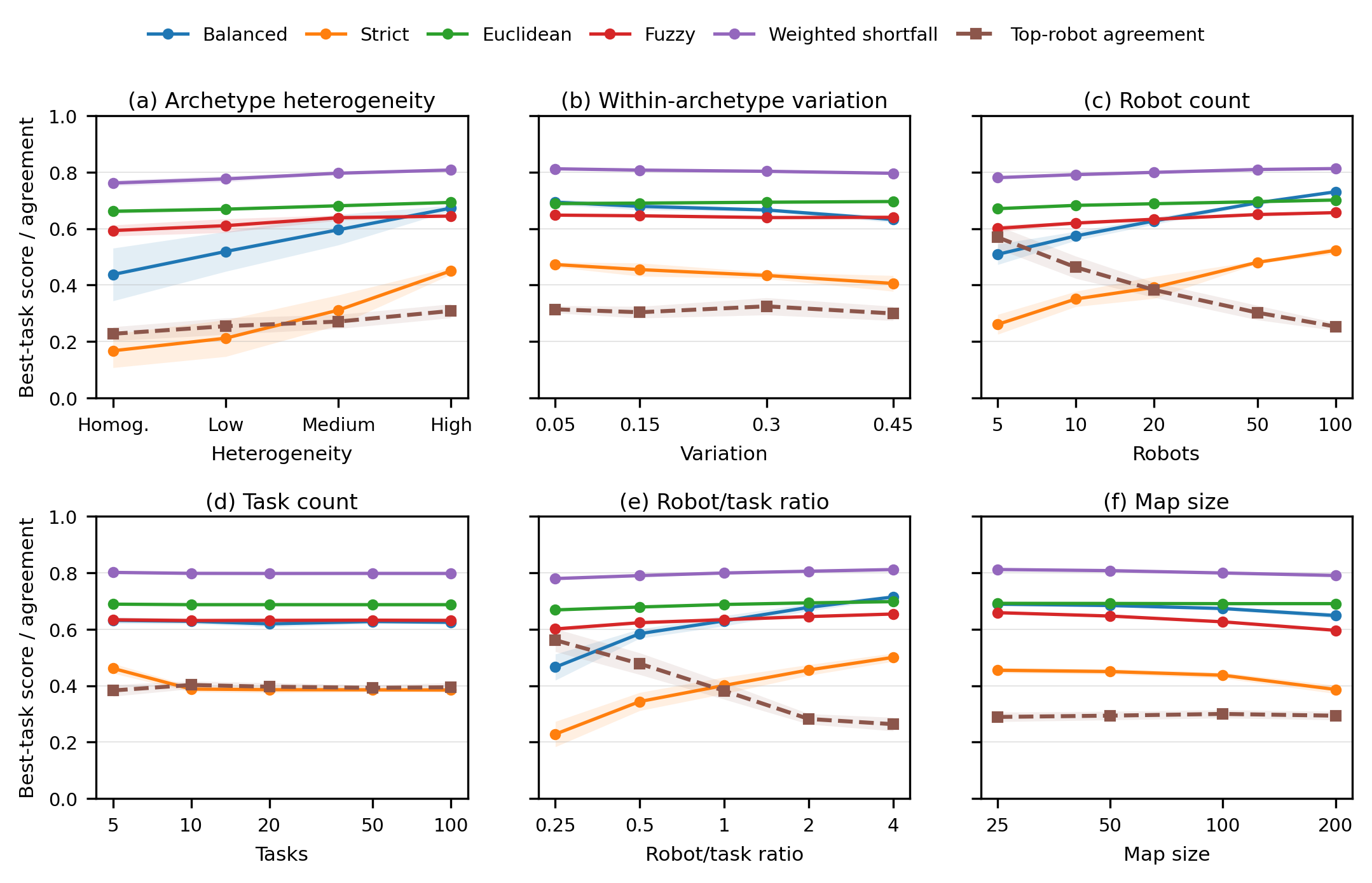}
    \caption{Effects of (a) between-archetype heterogeneity,
    (b) within-archetype variation, (c) robot count, (d) task count,
    (e) robot-to-task ratio, and (f) map size. Colored curves show mean
    best-task suitability; the black curve
    shows mean cross-method top-robot agreement.}
    \label{fig:experimental_sweeps}
\end{figure}

We evaluated Gemini 3.5 Flash-Lite as an LLM-based semantic suitability
estimator. Unlike the other
methods, Gemini is not given robot--task locations or the complete structured
task requirements. Its robot input contains the capability profile excluding
location, while its task input contains task type, priority, difficulty, and
natural-language description, from which requirements must be inferred. The
\textit{scalar} method returns one overall score; the \textit{structured} method evaluates
multiple capability dimensions before integrating them. Repeatability was
measured on a $10\times10$ scenario over ten identical evaluations using
entry-wise variability, pairwise score difference and correlation, rank
agreement, and top-robot agreement. Semantic robustness was evaluated by
replacing each task description with a paraphrase while
holding other attributes fixed. Gemini was compared with the
deterministic methods using the same score- and ranking-level measures.
Because the LLM receives a different task representation, these comparisons
reflect both the scoring mechanism and the available information.

\begin{table}[t]
\centering
\caption{Robustness to $20\%$ capability perturbation}
\label{tab:robustness_properties}
\scriptsize
\setlength{\tabcolsep}{4pt}
\begin{tabular}{lrrr}
\hline
Method &
\shortstack{Mean absolute\\score change} &
\shortstack{Score\\correlation} &
\shortstack{Top-robot\\stability (\%)} \\
\hline
Loose              & .0098 & .983 & 86.4 \\
Balanced           & .0181 & .937 & 85.6 \\
Strict             & .0055 & .953 & 90.2 \\
Euclidean          & .0028 & .984 & 65.0 \\
Cosine             & .0119 & .977 & 68.7 \\
Jaccard            & .0088 & .982 & 71.9 \\
Coverage           & .0169 & .957 & 68.0 \\
Fuzzy              & .0122 & .973 & 65.0 \\
Shortfall          & .0031 & .978 & 75.8 \\
Weighted shortfall & .0038 & .981 & 73.2 \\
\hline
\end{tabular}
\end{table}

\begin{table}[t]
\centering
\caption{Task-dependent deterministic agreement}
\label{tab:task_results}
\scriptsize
\setlength{\tabcolsep}{3.0pt}
\begin{tabular}{lrrrr}
\hline
Task type &
\shortstack{Mean absolute\\error} &
\shortstack{Mean per-task\\Spearman $\rho$} &
\shortstack{Exact top-robot\\agreement (\%)} &
\shortstack{Strict zero-\\valued pairs (\%)} \\
\hline

Assembly
& .252 & .711 & 37.3 & 100.0 \\

Debris
& .271 & .589 & 25.5 & 93.6 \\

Delivery
& .224 & .568 & 37.6 & 52.0 \\

Excavate
& .287 & .533 & 34.1 & 94.3 \\

Item elevation
& .267 & .705 & 43.2 & 86.9 \\

Lay bricks
& .240 & .428 & 18.5 & 76.9 \\

Scaffold
& .268 & .692 & 35.3 & 99.8 \\

Utilities
& .268 & .600 & 32.0 & 88.3 \\
\hline
\end{tabular}
\end{table}

\section{Results}
\label{sec:results}

The suitability methods differed substantially in score magnitude,
feasibility status, ranking, and top-robot selection. Strict and balanced
scoring produced many zero-valued pairs because of hard feasibility
conditions, whereas shortfall-based methods generally produced higher
nonzero scores; absolute suitability values should therefore not be
interpreted on a common calibrated scale. Figure~\ref{fig:experimental_sweeps}
summarizes the parameter sweeps. Increasing between-archetype heterogeneity
improved the best available robot--task match, whereas increasing
within-archetype variation generally reduced suitability and cross-method
agreement. Increasing the number of robots similarly improved the best
available match but reduced agreement on the top-ranked robot, while
increasing task count alone had little comparable effect. The robot-to-task
ratio sweep showed the same candidate-abundance effect, and increasing map
size primarily degraded context-sensitive methods while Euclidean and
Jaccard scores were comparatively invariant.

Table~\ref{tab:robustness_properties} shows that robustness of suitability
scores did not imply robustness of top-robot selection. Euclidean
exhibited the smallest mean score change and highest score correlation under
$20\%$ perturbation, but retained the same top robot for only $65.0\%$ of
tasks, compared with $90.2\%$ for strict scoring. Feasibility changes were
rare: zero/nonzero status flipped for $3.45\%$ of balanced pairs and $0.99\%$
of strict pairs, with no flips for the remaining methods. Property tests
likewise exposed systematic differences between generic similarity and
requirement-oriented methods. All methods passed critical-capability removal,
dominance, and numeric monotonicity tests. Euclidean, cosine, and Jaccard
violated irrelevant-capability invariance in all tested cases, while Euclidean
and cosine violated requirement relaxation in $98.55\%$ of cases; no property
violations were observed for the rule-based, coverage, fuzzy, or shortfall
methods.

\begin{table}[t]
\centering
\caption{Gemini repeatability, semantic robustness, and deterministic-method agreement}
\label{tab:gemini_results}
\scriptsize
\setlength{\tabcolsep}{2pt}
\begin{tabular}{p{66pt}rrrrr}
\hline
Comparison &
\shortstack{Mean entry\\std. dev.} &
\shortstack{Mean\\MAE} &
\shortstack{Matrix\\Pearson $r$} &
\shortstack{Mean task\\Spearman $\rho$} &
\shortstack{Top-robot\\agreement (\%)} \\
\hline

Scalar repeatability
& .074 & .082 & .882 & .840 & 85.3 \\

Structured repeatability
& .041 & .049 & .795 & .742 & 62.2 \\
\hline

\multicolumn{6}{c}{\textit{Original vs.\ Paraphrase}} \\
\hline
Scalar
& -- & .092 & .882 & .848 & 63.0 \\

Structured
& -- & .072 & .822 & .669 & 40.0 \\
\hline

\multicolumn{6}{c}{\textit{Structured Gemini vs.\ deterministic methods}} \\
\hline

Balanced
& -- & .444 & .494 & .590 & 25.0 \\

Fuzzy
& -- & .161 & .325 & .547 & 16.0 \\

Coverage
& -- & .132 & .365 & .540 & 26.0 \\

Shortfall
& -- & .116 & .382 & .439 & 11.0 \\

Strict
& -- & .586 & .295 & .409 & 26.0 \\

Weighted shortfall
& -- & .114 & .320 & .355 & 23.0 \\

Jaccard
& -- & .228 & .224 & .319 & 18.0 \\

Cosine
& -- & .181 & .169 & .307 & 10.0 \\

Loose
& -- & .272 & .186 & .228 & 9.0 \\

Euclidean
& -- & .111 & .038 & .031 & 10.0 \\
\hline
\end{tabular}
\end{table}

Agreement also depended strongly on task type
(Table~\ref{tab:task_results}). Item elevation produced the highest exact
top-robot agreement ($43.2\%$) and predominantly selected cranes, while
excavation similarly favored excavators across most methods. Brick laying
produced the lowest top-robot agreement ($18.5\%$) and the lowest mean
per-task rank correlation ($0.428$), with different methods favoring assembly
robots, bricklayers, or cranes. Strict feasibility varied substantially by
task: all strict assembly scores were zero and nearly all scaffold scores were
zero, so nominal top-ranked robots in those cases reflect all-zero ties rather
than meaningful preferences. Debris and utilities each combine gripper- and bucket-based task variants, so their agreement values aggregate different manipulation requirements. Overall, tasks with specific requirements produced more consistent suitability judgments than
tasks with several plausible capability combinations.

Gemini exhibited substantial run-to-run variability despite identical inputs
(Table~\ref{tab:gemini_results}). Structured prompting reduced score-level
variation relative to direct scalar scoring: mean entry-wise standard
deviation decreased from $0.074$ to $0.041$, maximum entry-wise standard
deviation from $0.320$ to $0.092$, and mean pairwise absolute error from
$0.082$ to $0.049$. This reduction did not extend to ranking or top-choice
stability: scalar repetitions had higher mean pairwise matrix correlation
($0.882$ versus $0.795$), mean per-task Spearman correlation
($0.840$ versus $0.742$), and exact top-robot agreement
($85.3\%$ versus $62.2\%$). However, scalar scoring produced substantially
more tied maxima ($64\%$ of task--repetition cases versus $24\%$ for
structured scoring), so deterministic tie-breaking likely inflates its
apparent top-choice stability; only $20\%$ of tasks retained a unanimous
structured top robot.

Semantic robustness showed the same distinction between score-level and
decision-level stability. When only the natural-language task descriptions
were paraphrased and all structured task attributes were held fixed, scalar
scoring produced a mean absolute difference of $0.092$, matrix-level Pearson
correlation of $0.882$, mean per-task Spearman correlation of $0.848$, and
$63.0\%$ exact top-robot agreement, while structured scoring produced a
smaller absolute difference ($0.072$) but lower matrix correlation ($0.822$),
rank correlation ($0.669$), and top-robot agreement ($40.0\%$). Scalar
scoring again produced more tied maxima, so its higher apparent decision
stability should be interpreted cautiously. Gemini also did not reproduce any
single deterministic method: its strongest mean per-task rank agreement
was with balanced scoring ($\rho=0.590$), followed by fuzzy ($\rho=0.547$)
and coverage ($\rho=0.540$), all of which
emphasize satisfaction of task requirements. Agreement was substantially
weaker with generic similarity measures, with Euclidean similarity showing essentially no
per-task rank agreement ($\rho=0.031$) despite having the
smallest mean absolute difference from structured scoring ($0.111$). This further shows that similar score magnitudes do not imply similar robot rankings.

\section{Discussion}
\label{sec:discussion}

The central implication of these experiments is that suitability methods
encode meaningfully different notions of robot--task compatibility. Rule-based
and requirement-oriented methods emphasize whether task requirements are
satisfied, whereas Euclidean, cosine, and Jaccard characterize broader
similarity between robot and task representations. The property tests make
this distinction explicit: Euclidean, cosine, and Jaccard penalized
task-irrelevant added capabilities, while Euclidean and cosine frequently
reduced suitability when task requirements were relaxed. These behaviors are
reasonable for symmetric similarity measures but conflict with a one-sided
requirement-satisfaction interpretation in which additional capability should
not hurt and an easier task should not become less suitable. The rule-based,
coverage, fuzzy, and shortfall methods satisfied all tested properties,
although this does not establish that they better predict execution outcomes.
Suitability measures should therefore be selected according to the semantics
and behavioral properties required by the allocation system rather than
solely for mathematical convenience or numerical agreement.

The heterogeneity and task-specific experiments further show that the
consequences of method choice depend on team composition and candidate
availability. Adding distinct robot archetypes increased the best available
estimated suitability and often made specialist relationships more
pronounced, whereas increasing variation within an archetype generally reduced
compatibility and cross-method agreement. Increasing the number of robots
similarly improved the best available match while reducing agreement about
which individual robot should be preferred, suggesting that method
choice becomes more consequential as several plausible candidates compete
under different capability tradeoffs. Task-level results support this
interpretation: agreement was stronger for tasks associated with distinctive
specialists and weaker for tasks, such as brick laying, that admit multiple
plausible capability strategies. Thus, heterogeneity can both improve the
quality of available matches and increase ambiguity about how those matches
should be ranked.

The spatial and perturbation experiments motivate a distinction between
\emph{intrinsic capability compatibility} and \emph{contextual suitability}.
Methods incorporating distance, endurance, sensing, or navigation changed
substantially with map scale, whereas more static capability-similarity
measures changed little. A robot may therefore possess the hardware needed for
a task while still being a poor current candidate because of location or
operating conditions, suggesting value in separating relatively static
compatibility from dynamic feasibility or execution cost. The perturbation results highlight the importance of evaluating suitability at the level at which it will be used: small score changes can still alter the preferred robot when candidates are closely matched, while larger score changes may leave the allocation decision unchanged. Because suitability is
ultimately used to support allocation decisions, ranking margins, top-choice
stability, and feasibility changes may be more informative than score
correlation alone.

LLM-based suitability introduces additional uncertainty from both stochastic
inference and linguistic representation. Structured prompting reduced
score-level variability relative to direct scalar judgments, suggesting that
decomposing compatibility into explicit capability dimensions can regularize
the generated values, but it did not improve ranking or top-choice stability.
Scalar scoring showed higher apparent decision stability, although its much
larger number of tied maxima makes this comparison difficult to interpret.
Paraphrasing otherwise equivalent task descriptions also changed
allocation-relevant rankings, demonstrating that linguistic representation
becomes part of the suitability function. Gemini did not
reproduce any deterministic method, but its rankings agreed more strongly with requirement-oriented balanced, fuzzy, and coverage scoring than with generic geometric similarity measures. Although this does not reveal the model's internal reasoning, it suggests Gemini’s suitability judgments align more with inferred requirement satisfaction than feature-space similarity. LLM-generated
suitability should therefore be treated as a distinct and uncertain
representation rather than as a deterministic approximation of conventional
capability matching; operational use may require standardized descriptions,
repeated sampling, aggregation, or allocation methods that explicitly account
for uncertainty.

These conclusions are limited by the absence of execution-derived ground
truth. The experiments characterize agreement, robustness, and behavioral
properties rather than predictive accuracy, so higher estimated suitability
cannot be interpreted as improved real-world task performance. The robot and
task archetypes are manually constructed and cover a finite set of capability
combinations, while the decision-level analysis considers top-robot selection
rather than complete allocations involving contention, coalitions, scheduling,
or other coupled constraints. Hard-feasibility methods also require special
treatment when every candidate receives zero suitability, since a nominal
$\arg\max$ then does not represent a meaningful feasible choice. Future work
should evaluate these methods against execution outcomes such as success,
completion time, energy use, and safety; expand the range and scale of robot
and task profiles; and measure how suitability disagreement propagates into
complete MRTA solutions. Overall, robot--task suitability should be
treated as an explicit modeling choice whose assumptions, context, and uncertainty are
explicit and matched to the needs of the allocation system.

\bibliographystyle{ieeetr}
\bibliography{mypapers,references}

\end{document}